\documentclass{iopconfser}

\usepackage[backend=biber,style=numeric,backref=true]{biblatex}

\newcommand{\figref}[1]{Figure~\ref{fig:#1}}

\renewcommand{\eqref}[1]{Eq.~(\ref{eq:#1})}

\newcommand{\edit}[1]{\textcolor{black}{{#1}}}

\newcommand{\deitp} {$\Delta$E$_\text{ITP}$}

\begin{document}

\title{A Neural Quality Metric for BRDF Models}

%\author{paper\_14}

\author{Behnaz Kavoosighafi$^{1}$, Rafał K. Mantiuk$^{2}$, Saghi Hajisharif$^{1}$, Ehsan Miandji$^{1}$, and Jonas Unger$^{1}$}

\affil{$^1$Department of Science and Technology, Linköping University, Norrköping, Sweden}

\affil{$^2$Department of Computer Science and Technology, University of Cambridge, Cambridge, United Kingdom}

\email{behnaz.kavoosighafi@liu.se}

\begin{abstract}
Accurately evaluating the quality of bidirectional reflectance distribution function (BRDF) models is essential for photo-realistic rendering. Traditional BRDF-space metrics often employ numerical error measures that fail to capture perceptual differences evident in rendered images. In this paper, we introduce the first perceptually informed neural quality metric for BRDF evaluation that operates directly in BRDF space, eliminating the need for rendering during quality assessment. Our metric is implemented as a compact multi-layer perceptron (MLP), trained on a dataset of measured BRDFs supplemented with synthetically generated data and labelled using a perceptually validated image-space metric. The network takes as input paired samples of reference and approximated BRDFs and predicts their perceptual quality in terms of just-objectionable-difference (JOD) scores. We show that our neural metric achieves significantly higher correlation with human judgments than existing BRDF-space metrics. While its performance as a loss function for BRDF fitting remains limited, the proposed metric offers a perceptually grounded alternative for evaluating BRDF models.
\end{abstract}

\section{Introduction}
Material appearance is a key component of photo-realistic rendering and is influenced by several factors, including surface reflectance, illumination, and geometry. The bidirectional reflectance distribution function (BRDF) describes surface reflectance by defining how incident light is reflected in different outgoing directions at a given wavelength~\cite{Nicodemus1977}. Since accurately modelling BRDFs with analytical functions often requires substantial domain expertise and careful parameter tuning, measured BRDF datasets~\cite{Matusik2003, Dupuy2018Adaptive} have become an essential resource for capturing complex reflectance behaviour without relying on simplified assumptions or isolated perceptual attributes. 

Rendering with measured BRDFs often requires approximating them using analytical or learned models. This fitting process typically relies on basic cost functions, such as $\ell_1$ or $\ell_2$ distances computed in either linear or logarithmic reflectance space. A fundamental property of these metrics is that they operate in the BRDF representation space, treating the reflectance function independently of any scene context, such as geometry or lighting. While this simplifies optimisation, it can lead to fitted models that are numerically accurate but visually inconsistent with the original material, resulting in perceptual artifacts in the final renderings. The mismatch between BRDF-space and image-space quality metrics has been previously observed, but partially addressed, in prior work \cite{Bieron2020, Kavoosighafi2025}.
For instance, some works~\cite{Lissner2013, Bieron2020, Havran2016} have explored evaluating the quality of BRDF fits using image-space metrics, such as PSNR, \deitp{}, and SSIM, computed over rendered outputs. These metrics inherently capture the influence of scene-dependent factors and have been shown in prior studies~\cite{Bieron2020, Kavoosighafi2025} to align more closely with human perception. However, applying these metrics during optimisation is computationally expensive, as it requires rendering the BRDF at each step. This makes them impractical for iterative fitting in most workflows.

To address this limitation, we propose a novel perceptually-based metric for evaluating the quality of BRDF model fits. While operating in the BRDF space, the metric takes scene information into account by learning a mapping from the input BRDFs to the output perceived quality in image space. We use the result of a perceptual study on a stereoscopic high dynamic range (HDR) display to train our neural model. Our evaluation shows that our neural metric correlates with the perceptual quality of rendered images, as assessed by human observers. We conduct evaluations to demonstrate the effectiveness of our proposed metric, comparing it to traditional BRDF-space metrics in terms of correlation with subjective data. 

Our key contributions are:
\begin{itemize}
    \item First neural quality metric for BRDF evaluation that operates directly in the BRDF space and does not require rendered images. The full source code for our metric is available at \href{https://github.com/behnazkavoosi/neural-brdf-quality-metric}{https://github.com/ behnazkavoosi/neural-brdf-quality-metric}.
    \item Significantly higher correlation with human judgments than existing BRDF-space metrics, providing a more perceptually reliable evaluation tool.
    \item Compact and efficient MLP architecture trained on a balanced, augmented BRDF dataset with perceptually aligned pseudo-labels, enabling real-time quality prediction.
\end{itemize}

\section{Background}

Since the introduction of measured BRDF datasets aimed at enhancing material realism in rendering applications, the development of perceptually meaningful BRDF metrics has become a central research focus. These metrics play an important role in BRDF fitting and optimisation, as well as in evaluating the performance of BRDF modelling and sampling techniques. Broadly, such metrics are computed either in BRDF space, emphasising representation error, or in image space, where they assess rendering error to evaluate visual accuracy.

\subsubsection*{BRDF Space.}
A wide range of cost functions has been proposed for BRDF optimisation, with $\ell_1$ and $\ell_2$ error measures forming the foundation of many such metrics. These metrics are often weighted by the cosine of the incoming and outgoing directions to account for angular reflectance variations. To improve alignment with human perception, researchers have explored perceptually motivated formulations, such as cube-root and logarithmic transformations. Cube-root cosine-weighted metrics have been shown to better capture perceptual differences, particularly around specular highlights~\cite{Fores2012, Lavoue2021}. Logarithmic metrics, on the other hand, are designed to account for perceptual nonlinearity and to reduce sensitivity to grazing angles~\cite{Low2012, Clausen2018}. While these BRDF-space approaches are simple and provide robust, scene-independent numerical approximations, they often fail to capture perceptual and visual differences in rendered appearance~\cite{Bieron2020}. This limitation was further highlighted in a recent user study evaluating the perceived quality of BRDF models, where BRDF-space metrics demonstrated weak correlation with human judgments~\cite{Kavoosighafi2025}.

\subsubsection*{Image Space.}
To assess BRDF similarity from a perceptual standpoint, many studies have employed image-based metrics evaluated under controlled rendering conditions. Early approaches relied on pixel-wise losses such as the $\ell_2$ norm under fixed lighting environments~\cite{Ngan2005}, later improved through structural similarity measures applied to tone-mapped images~\cite{Zhou2004, Brady2014}. Subsequent work extended this by exploring a range of perceptual metrics, including colour difference formulas, contrast-based indices, and high dynamic range visibility models, to evaluate differences in appearance, particularly for complex reflectance behaviours like anisotropy~\cite{Havran2016, Lissner2013, mantiuk2011hdr}. More recent methods leverage similarity metrics and combine them with statistical loss functions to select perceptually plausible BRDF fits from numerically optimised candidates~\cite{Bieron2020}. Large-scale user studies have also been conducted to better understand material appearance across varying lighting and geometric setups, leading to the development of perceptual similarity metrics grounded in human judgments~\cite{Lagunas2019, Lavoue2021}. These efforts have been further expanded to include dynamic content, where user ratings of video sequences indicated that perceptual similarity can be effectively approximated using combinations of correlation-based and norm-based measures~\cite{Filip2024}, as well as through colour difference computations in the ICtCp colour space~\cite{Kavoosighafi2025}.

\section{Neural Quality Metric}
Our goal is to develop a neural quality metric that predicts rendering quality given two input BRDFs: one representing the reference and the other the approximation. To train this model in a supervised manner, a comprehensive dataset is required that includes both BRDFs in their representation space along with their corresponding renderings, enabling the computation of image-space error. Once trained, the model can be used to predict image-space error directly from tabulated BRDFs in the test set, without requiring additional renderings.

\subsection{Dataset and Pre-Processing}

We build on a dataset introduced in prior work~\cite{Kavoosighafi2025}, which contains rendered video sequences for 159 BRDF material samples sourced from the MERL~\cite{Matusik2003}, RGL-EPFL~\cite{Dupuy2018Adaptive}, and DTU~\cite{Nielsen2015} datasets. The material samples were used to render short video clips showing a rotating Stanford bunny illuminated with the Pixar Campus environment map. Each BRDF was approximated using nine different BRDF models at varying levels of encoding, resulting in a total of 1\,431 distorted videos. The dataset also includes subjective quality ratings in the form of just-objectionable-difference (JOD) scores. However, the perceptual study was conducted on a subset of only 20 BRDF samples, providing ground truth for 180 distorted instances. To enable supervised training across the full dataset, we generated pseudo-labels for the remaining data using an alternative image-space metric. 

\begin{figure}[t]
	\centering
	\setlength{\tabcolsep}{0.1cm}
	\begin{tabular}{cc}
    \begin{subfigure}[t]{0.49\textwidth}
			\centering
			\hspace{-20pt}\includegraphics[width=\linewidth,keepaspectratio]{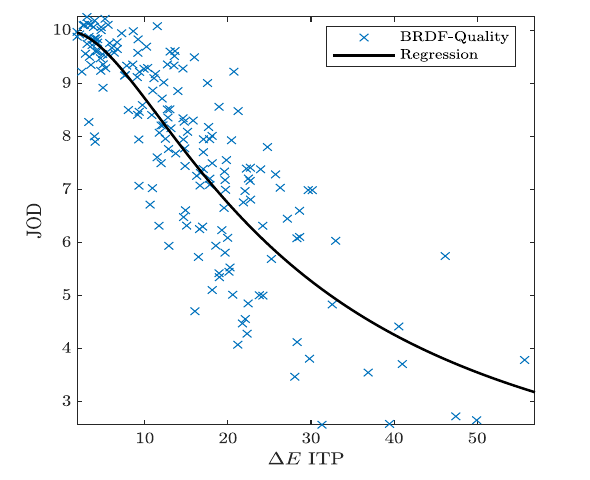}
			\caption{}
			\label{fig:jod-vs-de}
		\end{subfigure}
        &
		\begin{subfigure}[t]{0.49\textwidth}
			\centering
			\hspace{-40pt} \includegraphics[width=\linewidth,keepaspectratio]{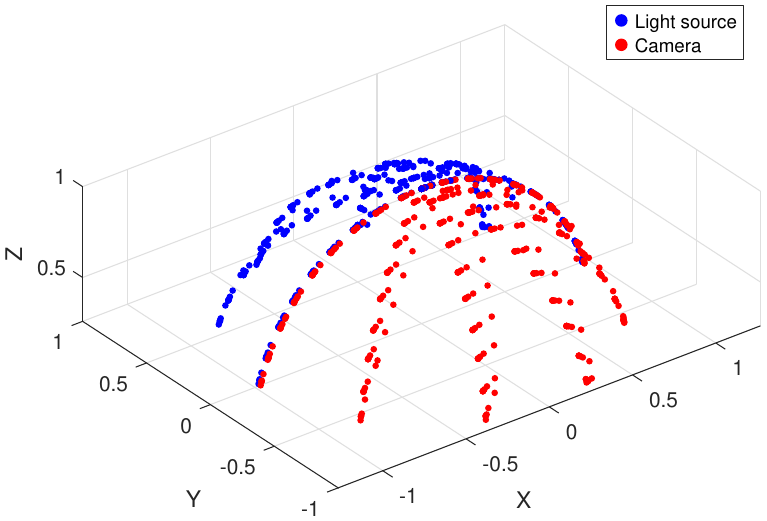}
			\caption{}
			\label{fig:directions}
		\end{subfigure}
		%&
		%\begin{subfigure}[t]{0.33\textwidth}
			%\centering
			%\hspace{-40pt}\includegraphics[width=\linewidth,keepaspectratio]{figs/balanced.pdf}
		%	\caption{}
		%	\label{fig:balanced}
		%\end{subfigure}
		
	\end{tabular}
	\caption{(a) Relationship between JOD values and \deitp{} in the original dataset, illustrating the correlation between subjective quality assessments and perceptual colour differences. \edit{(b) Sampled BRDF locations in the angular domain.}} %(b) Original dataset with an imbalanced JOD distribution skewed toward high-quality samples. (c) Augmented dataset with a more balanced JOD distribution across the quality spectrum. }
	\label{fig:jod-dist}
\end{figure}

In the previous study~\cite{Kavoosighafi2025}, the \deitp{} metric demonstrated the strongest correlation with human judgments among the evaluated metrics (see \figref{jod-vs-de}). We therefore used this \edit{image-space} metric to generate labels for training. For the subset of materials used in the perceptual experiment, where ground-truth JOD values are available, we fitted the metric predictions to the corresponding JOD scores using the Levenberg–Marquardt algorithm~\cite{Levenberg1944, Marquardt1963} (\verb!lsqcurvefit! function in MATLAB) to estimate the optimal regression parameters. Once the parameters were obtained, we used them to compute JOD values for all materials in the dataset based on their \deitp{} values:
\begin{equation}
\text{JOD} = 10 \cdot \left(1 - \frac{1}{1 + \exp\left(b_1 \cdot \left(-\left(\max(\text{\deitp{}}, 0)^{b_3}\right) - b_2\right)\right)}\right),
\label{eq:jod-regression}
\end{equation}
where $b_1 = -14.11$, $b_2 = -0.47$, and $b_3 = -0.21$ are the fitted parameters. 

To prepare the inputs for training the neural network, we reduced the angular complexity of the BRDFs by sampling a subset of values from valid angular regions. We extracted 500 distinct BRDF reflectance samples per material from the available measured datasets. This subsampling facilitates the use of simpler and more efficient network architectures while preserving key reflectance characteristics. All BRDFs were first transformed into the Rusinkiewicz coordinate system~\cite{Rusinkiewicz1998}. Samples were then selected based on their reflectance magnitude, prioritizing directions with significant energy. For the difference vector, we applied uniform sampling across both the elevation ($\theta_d$) and azimuthal ($\phi_d$) angles to ensure diverse angular coverage. For the half-vector elevation angle ($\theta_h$), we adopted a non-uniform sampling strategy to increase sample density near $\theta_h = 0$, thus capturing the structure of specular highlights more effectively (see \figref{directions}). To avoid instability due to measurement noise, we discarded grazing angles above 75\textdegree~prior to sampling~\cite{Ngan2005}. This pre-processing step resulted in a dataset of 159 reference BRDFs and 1\,431 approximated BRDFs, each represented by 500 samples in the Rusinkiewicz domain. Every BRDF instance was paired with a scalar JOD value, providing the target output for supervised training.

Although our sampling strategy ensures a diverse coverage of BRDF reflectance samples across the angular domain, we observed a notable imbalance in the distribution of quality scores: samples with high JOD values (corresponding to subtle or imperceptible differences) were dominant, while low JOD values (indicating strong perceptual degradation) were significantly under-represented. This imbalance poses a challenge for supervised learning, biasing the network toward specific error levels. To address this, we augmented the dataset by generating additional BRDF pairs that emphasise the low-JOD region. The number of generated samples was informed by the histogram of existing JOD values, such that under-represented regions receive a greater proportion of synthetic examples. We did this by introducing controlled random noise into the BRDF values, ensuring that the relative error between the reference and distorted BRDF remains consistent. We perturbed each BRDF's reflectance values using additive Gaussian noise drawn from a normal distribution. For each colour channel, the noisy BRDF was computed as
$\rho_{n} = \rho + \mathcal{N}(0, \sigma^2)$, 
where $\rho$ and $\rho_{\text{n}}$ denote the original and noisy BRDF values, respectively, and $\sigma^2$ is the variance of the noise. In our implementation, we set $\sigma = 0.01$ to ensure small but meaningful deviations that simulate low-quality distortions. \edit{The noise was added independently to each sampled direction without enforcing angular correlation. While angularly correlated alternatives could be explored in future work, this approach was effective for balancing the dataset and enhancing performance.} This augmentation step generated a total of 3\,340 unique BRDF pairs with a more uniform JOD distribution. %as shown in Figures \ref{fig:unbalanced} and  \ref{fig:balanced}. 

The dataset was randomly split into training (80\%) and validation (20\%) sets, while the BRDFs used in the perceptual experiment were reserved for testing. This resulted in 2\,672 BRDF pairs for training, 668 for validation, and 180 for testing. To further enrich the training set, we applied a second augmentation step in which each pair of test and reference BRDFs was scaled by a random factor sampled uniformly from the range $[0.95, 1.05]$, producing a final training set of 5\,344 BRDF pairs.

We standardised the reflectance values through a series of transformations designed to enhance learning and align the input representation with perceptual characteristics. First, a cube root transformation was applied to suppress sharp specular peaks, which are often overemphasised in raw reflectance values~\cite{Lavoue2021}. \edit{Next, a logarithmic transformation was used to compress the dynamic range and account for the non-linear response of human vision~\cite{Low2012}, defined as:
\begin{equation}
\rho_t = \log\left(\rho^{1/3} + 1\right),
\end{equation}
where $\rho$ denotes the raw reflectance value and $\rho_t$ is the transformed value.} Finally, we applied whitening by subtracting the mean and dividing by the standard deviation of each channel, computed over the reference BRDFs in the training set. All transformations were applied identically to both reference and distorted BRDFs to preserve relative differences.

\begin{figure}[t] 
	\centering
	\setlength{\tabcolsep}{0cm}
	\includegraphics[width=0.8\linewidth]{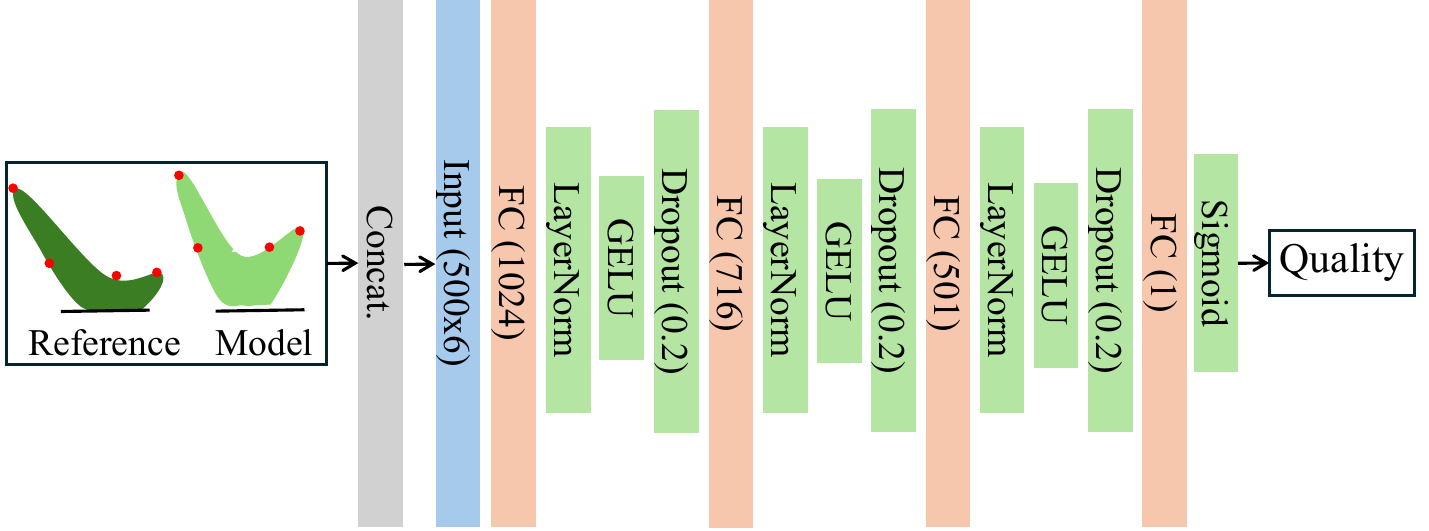}
	\caption{Network architecture: The input layer is the concatenation of reference and model RGB BRDFs with 500 samples, four samples are illustrated with red dots, leading to a vector of size $500\times6$. The output is a scalar corresponding to JOD value. Each fully connected layer (FC), except the last, is followed by layer normalisation, Gaussian Error Linear Unit (GELU) activation, and a dropout layer with a probability of 0.2.}
	\label{fig:network}
\end{figure}

\subsection{Network Architecture}

Given the ability of multi-layer perceptrons (MLPs) to approximate complex non-linear functions, we design an MLP-based regressor to predict perceptual quality scores (JOD values) from paired reference and approximated BRDF samples (see \figref{network}). The MLP architecture comprises four fully connected layers. Each hidden layer is followed by Layer Normalisation~\cite{Ba2016} to stabilise training dynamics, a GELU (Gaussian Error Linear Unit) activation~\cite{Hendrycks2016} to introduce smooth non-linearity, and a dropout layer with a rate of 0.2 to mitigate overfitting. To ensure that the predicted JOD values lie within the expected perceptual scale, a sigmoid activation is applied after the final layer, re-scaling the output to the range defined by the minimum and maximum JOD values observed in the dataset.

The input to the network is a concatenation of the reference and distorted BRDF samples, each of dimension $500 \times 3$, resulting in an overall input shape of $500 \times 6$. The first linear layer expands the representation to 1\,024 features. The subsequent layers reduce the dimensionality progressively using a decay factor of approximately 0.7, resulting in 716 and 501 features in the second and third layers, respectively. The final output layer produces a single scalar value corresponding to the predicted JOD score. To enhance robustness to outliers while maintaining differentiability and stable gradients, we used the log-cosh loss for training, which is defined as:
\begin{equation}
l = \log\left(\cosh(x - y)\right),
\end{equation}
where $x$ is the predicted JOD value and $y$ is the corresponding ground truth. Training was conducted for 100 epochs using the Adam optimiser~\cite{Kingma2014}, with a base learning rate of $10^{-4}$ for the input layer and $10^{-3}$ for the deeper layers. We applied a weight decay of $10^{-4}$ to enhance generalisation and used a batch size of 512. To adaptively reduce the learning rate when validation loss stagnates, we employed PyTorch’s \verb!ReduceLROnPlateau! scheduler with a patience of 5 epochs. On an RTX 4080 GPU, training took approximately 36 seconds, while inference required only 0.04 milliseconds per prediction. The final model comprises 4\,171\,125 trainable parameters and requires 15.91 MB of storage when saved in 32-bit floating-point format.

\begin{figure}[ht] 
	\centering
	\setlength{\tabcolsep}{0cm}
	\hspace{-10pt}\includegraphics[width=0.5\linewidth]{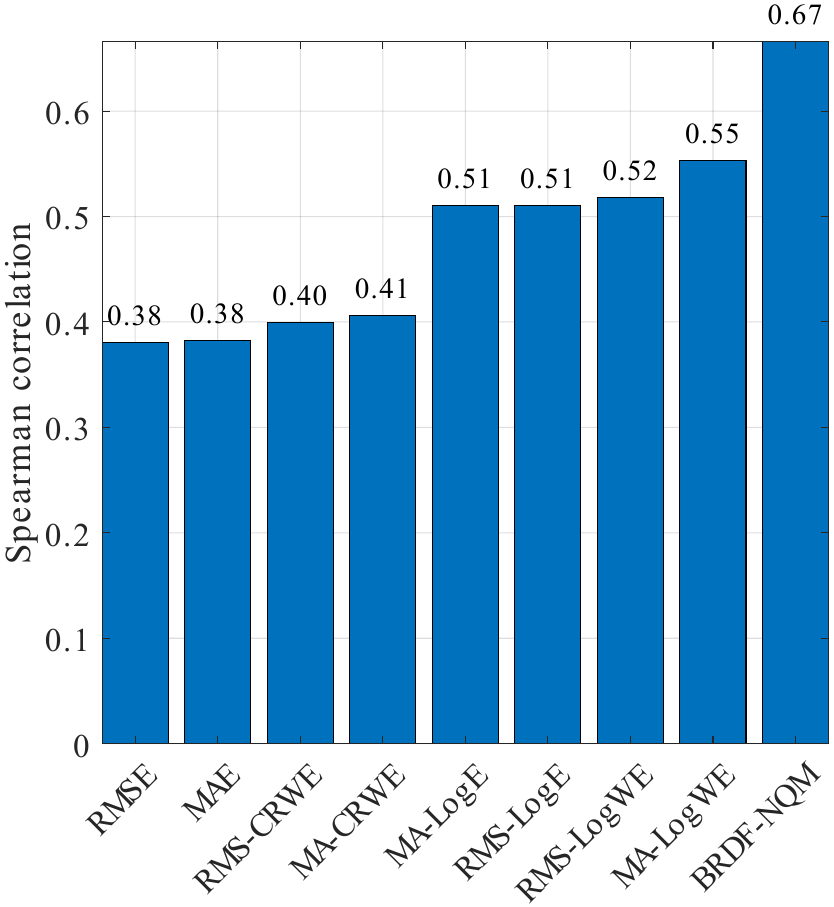}
	\caption{Average Spearman correlation between subjective JOD scores and various BRDF-space metrics. Our neural quality metric (BRDF-NQM) is included for comparison against traditional BRDF-space metrics. Higher values indicate stronger alignment with human perceptual judgments.}
	\label{fig:nqm-result}
\end{figure}

\section{Results}

We evaluate the performance of our neural quality metric by comparing its predictions with the ground-truth JOD values obtained from the user study conducted in previous work~\cite{Kavoosighafi2025}. As discussed, our model was trained using pseudo-labels derived from \deitp{} scores, which introduces some discrepancy between the training targets and the actual subjective data. Nevertheless, we compare our model against a set of BRDF-space metrics evaluated in the same prior study, selecting the best-performing variant of each metric to ensure a fair comparison.

The baseline metrics include root mean squared error (RMSE) and mean absolute error (MAE), which operate directly on linear BRDF values. In addition to these, we evaluate perceptually motivated variants that apply nonlinear transformations and weighting strategies. These include the root mean squared and mean absolute cubic root-weighted errors (RMS-CRWE and MA-CRWE), which apply a cube root transformation and cosine weighting to emphasise perceptually significant regions. Similarly, the root mean squared and mean absolute logarithmic errors (RMS-LogE and MA-LogE) use a logarithmic transformation to compress the dynamic range in line with perceptual nonlinearity. Finally, root mean squared and mean absolute logarithmic weighted errors (RMS-LogWE and MA-LogWE) combine logarithmic transformation with cosine weighting to better model human sensitivity to directional reflectance variation.

As shown in \figref{nqm-result}, our proposed method, denoted as \emph{BRDF-NQM}, achieves the highest correlation with subjective JOD scores, reaching an average Spearman correlation of 0.67. A higher correlation value indicates stronger agreement with human perceptual judgments. This result demonstrates a significant improvement over all tested BRDF-space metrics and highlights the effectiveness of learning a perceptually grounded metric directly from BRDF data.

Perceptual metrics are often employed as loss functions in BRDF fitting pipelines, where they are expected to guide optimisation toward perceptually meaningful results. To investigate this use case, we applied our neural metric as a loss function for fitting measured BRDFs. While certain reflectance properties, such as the index of refraction and surface roughness, were predicted with reasonable accuracy, we consistently observed a colour shift toward red in the estimated diffuse component. This artifact likely results from colour imbalances in the training data and reduced chromatic sensitivity introduced by the whitening step in pre-processing. These findings are consistent with prior work~\cite{Kavoosighafi2025}, which showed that a metric’s predictive accuracy does not necessarily translate into effective performance as a loss function. Bridging this gap remains an important direction for future research in perceptually guided BRDF modelling.

\section{Conclusion}

In this paper, we presented the first perceptually-informed neural quality metric for evaluating BRDFs directly in BRDF space. Designed to predict perceptual quality under complex viewing conditions, the metric employs a four-layer MLP that takes as input a pair of reference and approximated BRDFs, sampled to maintain the most perceptually significant reflectance information. We evaluated the proposed metric against the state-of-the-art BRDF-space metrics by measuring their correlation with human judgments and observed a significant improvement, establishing our method as the most effective BRDF-space metric for BRDF evaluation. While the metric shows strong predictive performance, it does not transfer as effectively when used as a stand-alone loss function in BRDF fitting. Nevertheless, our findings highlight the potential of perceptually grounded neural metrics to inform the design of new BRDF-space loss functions that prioritise perceptual fidelity, potentially reducing or eliminating the need for costly rendering in optimisation tasks.

\section*{Acknowledgments}
%*** Left blank to maintain anonymity for review.  *** 
This work is a part of PRIME, which is funded by the European Union’s Horizon 2020 research and innovation program under the Marie Skłodowska Curie grant agreement No 956585.

\printbibliography  

@article{Kavoosighafi2025,
journal = {Computer Graphics Forum},
title = {{Perceived Quality of BRDF Models}},
author = {Kavoosighafi, Behnaz and Mantiuk, Rafal K. and Hajisharif, Saghi and Miandji, Ehsan and Unger, Jonas},
year = {2025},
publisher = {The Eurographics Association and John Wiley & Sons Ltd.},
ISSN = {1467-8659},
DOI = {10.1111/cgf.70162}
}

@article{Fores2012,
author = {Forés, Adrià and Ferwerda, James and Gu, Jinwei},
year = {2012},
month = {01},
pages = {142-148},
title = {Toward a Perceptually Based Metric for BRDF Modeling},
volume = {CIC’12},
journal = {Final Program and Proceedings - IS and T/SID Color Imaging Conference},
doi = {10.2352/CIC.2012.20.1.art00025}
}

@article{Lavoue2021,
author = {Lavoué, Guillaume and Bonneel, Nicolas and Farrugia, Jean-Philippe and Soler, Cyril},
title = {Perceptual quality of BRDF approximations: dataset and metrics},
journal = {Computer Graphics Forum},
volume = {40},
number = {2},
pages = {327-338},
doi = {10.1111/cgf.142636},
year = {2021}
}

@article{Clausen2018,
author = {Clausen, O. and Marroquim, R. and Fuhrmann, A.},
title = {Acquisition and Validation of Spectral Ground Truth Data for Predictive Rendering of Rough Surfaces},
journal = {Computer Graphics Forum},
volume = {37},
number = {4},
pages = {1-12},
doi = {10.1111/cgf.13470},
year = {2018}
}

@article{Low2012,
author = {L\"{o}w, Joakim and Kronander, Joel and Ynnerman, Anders and Unger, Jonas},
title = {BRDF models for accurate and efficient rendering of glossy surfaces},
year = {2012},
issue_date = {January 2012},
publisher = {Association for Computing Machinery},
address = {New York, NY, USA},
volume = {31},
number = {1},
issn = {0730-0301},
doi = {10.1145/2077341.2077350},
journal = {ACM Trans. Graph.},
month = feb,
articleno = {9},
numpages = {14}
}

@inproceedings{Ngan2005,
booktitle = {Eurographics Symposium on Rendering (2005)},
editor = {Kavita Bala and Philip Dutre},
title = {{Experimental Analysis of BRDF Models}},
author = {Ngan, Addy and Durand, Frédo and Matusik, Wojciech},
year = {2005},
publisher = {The Eurographics Association},
ISSN = {1727-3463},
ISBN = {3-905673-23-1},
DOI = {10.2312/EGWR/EGSR05/117-126}
}

@article{Brady2014,
author = {Brady, Adam and Lawrence, Jason and Peers, Pieter and Weimer, Westley},
title = {genBRDF: discovering new analytic BRDFs with genetic programming},
year = {2014},
issue_date = {July 2014},
publisher = {Association for Computing Machinery},
address = {New York, NY, USA},
volume = {33},
number = {4},
issn = {0730-0301},
doi = {10.1145/2601097.2601193},
journal = {ACM Trans. Graph.},
month = jul,
articleno = {114},
numpages = {11}
}

@ARTICLE{Zhou2004,
  author={Zhou Wang and Bovik, A.C. and Sheikh, H.R. and Simoncelli, E.P.},
  journal={IEEE Transactions on Image Processing}, 
  title={Image quality assessment: from error visibility to structural similarity}, 
  year={2004},
  volume={13},
  number={4},
  pages={600-612},
  doi={10.1109/TIP.2003.819861}
}

@article{Bieron2020,
author = {Bieron, J. and Peers, P.},
title = {An Adaptive BRDF Fitting Metric},
journal = {Computer Graphics Forum},
volume = {39},
number = {4},
pages = {59-74},
doi = {10.1111/cgf.14054},
year = {2020}
}

@article{Lagunas2019,
author = {Lagunas, Manuel and Malpica, Sandra and Serrano, Ana and Garces, Elena and Gutierrez, Diego and Masia, Belen},
title = {A similarity measure for material appearance},
year = {2019},
issue_date = {August 2019},
publisher = {Association for Computing Machinery},
address = {New York, NY, USA},
volume = {38},
number = {4},
issn = {0730-0301},
doi = {10.1145/3306346.3323036},
journal = {ACM Trans. Graph.},
month = jul,
articleno = {135},
numpages = {12}
}

@misc{Filip2024,
      title={Material Fingerprinting: Identifying and Predicting Perceptual Attributes of Material Appearance}, 
      author={Jiri Filip and Filip Dechterenko and Filipp Schmidt and Jiri Lukavsky and Veronika Vilimovska and Jan Kotera and Roland W. Fleming},
      year={2024},
      eprint={2410.13615},
      archivePrefix={arXiv},
      primaryClass={cs.CV} 
}

@article{mantiuk2011hdr,
  title={{HDR-VDP-2}: A calibrated visual metric for visibility and quality predictions in all luminance conditions},
  author={Mantiuk, Rafa{\l} and Kim, Kil Joong and Rempel, Allan G and Heidrich, Wolfgang},
  journal=TOG,
  volume={30},
  number={4},
  pages={1--14},
  year={2011},
  publisher={ACM New York, NY, USA}
}

@article{Havran2016,
author = {Havran, V. and Filip, J. and Myszkowski, K.},
title = {Perceptually Motivated BRDF Comparison using Single Image},
journal = {Computer Graphics Forum},
volume = {35},
number = {4},
pages = {1-12},
doi = {10.1111/cgf.12944},
year = {2016}
}

@ARTICLE{Lissner2013,
  author={Lissner, Ingmar and Preiss, Jens and Urban, Philipp and Lichtenauer, Matthias Scheller and Zolliker, Peter},
  journal={IEEE Transactions on Image Processing}, 
  title={Image-Difference Prediction: From Grayscale to Color}, 
  year={2013},
  volume={22},
  number={2},
  pages={435-446},
  doi={10.1109/TIP.2012.2216279}
}

@article{Levenberg1944,
    title={A method for the solution of certain nonlinear problems in least squares},
    author={Levenberg, Kenneth},
    journal={Quarterly of Applied Mathematics},
    volume={2},
    number={2},
    pages={164--168},
    year={1944}
}

@article{Marquardt1963,
    title={An algorithm for least-squares estimation of nonlinear parameters},
    author={Marquardt, Donald W.},
    journal={Journal of the Society for Industrial and Applied Mathematics},
    volume={11},
    number={2},
    pages={431--441},
    year={1963},
    publisher={SIAM}
}

@InProceedings{Rusinkiewicz1998,
author="Rusinkiewicz, Szymon M.",
editor="Drettakis, George
and Max, Nelson",
title="A New Change of Variables for Efficient BRDF Representation",
booktitle="Rendering Techniques '98",
year="1998",
publisher="Springer Vienna",
address="Vienna",
pages="11--22",
isbn="978-3-7091-6453-2"
}

@article{Matusik2003,
author = {Matusik, Wojciech and Pfister, Hanspeter and Brand, Matt and McMillan, Leonard},
title = {A Data-Driven Reflectance Model},
year = {2003},
issue_date = {July 2003},
publisher = {Association for Computing Machinery},
address = {New York, NY, USA},
volume = {22},
number = {3},
issn = {0730-0301},
doi = {10.1145/882262.882343},
journal = {ACM Trans. Graph.},
month = {07},
pages = {759–769},
numpages = {11}
}

@article{Dupuy2018Adaptive,
    author = {Jonathan Dupuy and Wenzel Jakob},
    title = {An Adaptive Parameterization for Efficient Material Acquisition and Rendering},
    journal = {Transactions on Graphics (Proceedings of SIGGRAPH Asia)},
    volume = {37},
    number = {6},
    pages = {274:1--274:18},
    year = {2018},
    month = nov,
    doi = {10.1145/3272127.3275059}
}

@article {Nielsen2015,
	author = {Jannik Boll Nielsen and Henrik Wann Jensen and Ravi Ramamoorthi},
	title = {On Optimal, Minimal BRDF Sampling for Reflectance Acquisition},
	journal = {ACM Transactions on Graphics (TOG)},
	year = {2015},
	month = {November},
	volume = {34},
	number = {6},
	pages = {186:1-186:11},
	doi = {10.1145/2816795.2818085}
}

@article{Kingma2014,
author = {Kingma, Diederik and Ba, Jimmy},
year = {2014},
month = {12},
pages = {},
title = {Adam: A Method for Stochastic Optimization},
journal = {International Conference on Learning Representations}
}

@article{Hendrycks2016,
  title={Gaussian Error Linear Units (GELUs)},
  author={Hendrycks, Dan and Gimpel, Kevin},
  journal={arXiv preprint arXiv:1606.08415},
  year={2016}
}

@article{Ba2016,
    author = "Ba, Jimmy Lei and Kiros, Jamie Ryan and Hinton, Geoffrey E.",
    title = "{Layer Normalization}",
    eprint = "1607.06450",
    archivePrefix = "arXiv",
    primaryClass = "stat.ML",
    month = "7",
    year = "2016"
}

@inproceedings{Nicodemus1977,
    title={{Geometrical Considerations and Nomenclature for Reflectance}},
    author={Fred E. Nicodemus and Joseph C. Richmond and Jack J. Hsia and Irving W. Ginsberg and T. Limperis and Sidney Harman and Jordan J Baruch},
    year={1977}
}

\end{document}